\documentclass[letterpaper, conference]{ieeeconf}

\usepackage{times}
\IEEEoverridecommandlockouts                              % This command is only needed if 

\usepackage{multicol}
\usepackage[bookmarks=true]{hyperref}
\usepackage{amsmath}
\usepackage{framed}
\usepackage{amssymb}
\usepackage{bm}
\usepackage[version-1-compatibility]{siunitx}

\usepackage{booktabs}
\usepackage{amsthm}
\usepackage{graphicx}
\usepackage[dvipsnames, table]{xcolor}
\usepackage{tikz}
\usepackage{blindtext}

\usepackage{amsmath}

\usepackage{pifont}% http://ctan.org/pkg/pifont (for check, xmark)
\newcommand{\cmark}{\ding{51}}%
\newcommand{\xmark}{\ding{55}}%

\tikzstyle axes label=[font=\footnotesize, text width=3cm, text centered]

\usepackage{color}
\usepackage{mathrsfs}

\graphicspath{{figures/}
  {../../medias/}
}

\theoremstyle{plain}

\theoremstyle{definition}
\newtheorem{definition2}{Definition}

\theoremstyle{definition}

\newtheorem{remark2}{Remark}

\newtheorem{assum2}{Assumption}
\newenvironment{assum}
  {%
   \pushQED{\qed}\begin{assum2}}
  {\popQED\end{assum2}}

\usepackage[ruled, linesnumbered]{algorithm2e}
\SetKwIF{If}{ElseIf}{Else}{if}{:}{elif}{else}{}%
\SetKwFor{For}{for}{ do}{}%
\SetKwFor{ForEach}{foreach}{ do}{}%
\SetKwInOut{Input}{Input}%
\SetKwInOut{Output}{Output}%
\SetKwInOut{Param}{Parameters}%
\SetAlgoNoLine%
\DontPrintSemicolon%
\SetKwFor{While}{while}{:}{}%
\SetKwFor{For}{for}{:}{}%

\DeclareFontFamily{U}{skulls}{}
\DeclareFontShape{U}{skulls}{m}{n}{ <-> skull }{}

\newcommand{\su}[1]{\textcolor{Green}{\textbf{#1}} \cmark}
\newcommand{\bsu}[1]{\textcolor{Green}{\textbf{#1}}}
\newcommand{\fa}[1]{\textcolor{Red}{#1} \xmark}

\renewcommand{\vec}[1]{\bm{\mathrm{#1}}}  % use bm to obtain boldface for symbols

\newcommand{\cC}{\mathscr{C}}

\begin{document}

% paper title
\title{\LARGE  Critically fast pick-and-place with suction cups}

\author{Hung Pham$^{*, \dagger}$, Quang-Cuong Pham$^{*}$
  \thanks{
    hungpham2511@gmail.com, cuong.pham@normalesup.org
    \newline \indent $^{*}$ Singapore Centre for 3D Printing (SC3DP), School of
    Mechanical and Aerospace Engineering, Nanyang Technological
    University, Singapore
    \newline \indent $^{\dagger}$ Eureka Robotics Pte. Ltd., Singapore}
}

\maketitle

\begin{abstract}
  Fast robotics pick-and-place with suction cups is a crucial component
  in the current development of automation in logistics (factory
  lines, e-commerce, etc.).  By ``critically fast'' we mean the
  fastest possible movement for transporting an object such that it
  does not slip or fall from the suction cup. The main difficulties
  are: (i) handling the contact between the suction cup and the
  object, which fundamentally involves kinodynamic constraints; and
  (ii) doing so at a low computational cost, typically a few hundreds
  of milliseconds. To address these difficulties, we propose (a) a
  model for suction cup contacts, (b) a procedure to identify the
  contact stability constraint based on that model, and (c) a pipeline
  to parameterize, in a time-optimal manner, arbitrary geometric paths
  under the identified contact stability constraint. We experimentally
  validate the proposed pipeline on a physical robot system: the cycle
  time for a typical pick-and-place task was less than 5 seconds,
  planning and execution times included. The full pipeline is released
  as open-source for the robotics community.
\end{abstract}

\IEEEpeerreviewmaketitle

\section{Introduction}
\label{sec:intro}

Suction cups have proved to be some of the most robust and versatile
devices to grasp a variety of objects, and are therefore used in a
large proportion of automation solutions in logistics (factory lines,
e-commerce, etc). In addition to robustness and versatility, cycle
time is another key driver in the development of automation: how fast
automated systems can pick and place objects is a decisive economic
question.

In this paper, we are interested in planning and executing critically
fast pick-and-place movements with suction cups. By ``critically
fast'' we mean the fastest possible movements for transporting an
object such that it does not slip or fall from the suction
cup. Another major concern is to also minimize the planning time.
Indeed, in a majority of e-commerce scenarios, the robot movements
must be computed based on the dynamically perceived positions of the
objects to be transported, which implies that both planning and
execution times contribute to the total cycle time.

Note that, to focus on the planning and execution aspects, we assume
in this paper that the perception problem has been solved upstream:
(i) a good model of the environment is available, (ii) the geometries,
weight distribution, and initial positions/orientations of the objects
are given and accurate.

Essentially, ensuring that the object being transported does not slip
or fall amounts to guaranteeing that, at every moment, the net
inertial wrench acting on the object can be physically ``realized'' by
its contact with the suction cup (weak contact
stability~\cite{pang2000stability,caron2015stability}). Since inertial
wrenches can be easily computed given a sufficiently accurate model of
the object, this task is reduced to identifying the set of wrenches
that are \emph{physically realizable} for the given suction cup
contact. A procedure for doing this is among the contributions of this
paper. In the following, we will refer to this objective as satisfying
the \emph{suction cup grasp stability constraint}, or simply
\emph{grasp stability constraint}.

In industrial robotic pick-and-place systems, a popular approach to
maintaining suction cup grasp stability is to uniformly restrict the
robot's joint velocity and acceleration. The velocity and acceleration
limits can be tuned by executing multiple test trajectories, then
selecting the set of limits at which there is no failure and the
average duration is the shortest. A drawback of this approach is that
testing must be done for each object/suction cup combination, and
therefore, is tedious and time consuming. Furthermore, it is certain
that the resulting movements are not optimal, since the kinematic
limits are chosen with respect to \emph{all} testing
trajectories. Also, there is no guarantee that the robot can
successfully execute any new, untested trajectory during its actual
operation.

\begin{figure}[t]
  \centering
  \includegraphics[trim={0cm 3cm 0cm 0cm}, clip, width=0.48\textwidth]{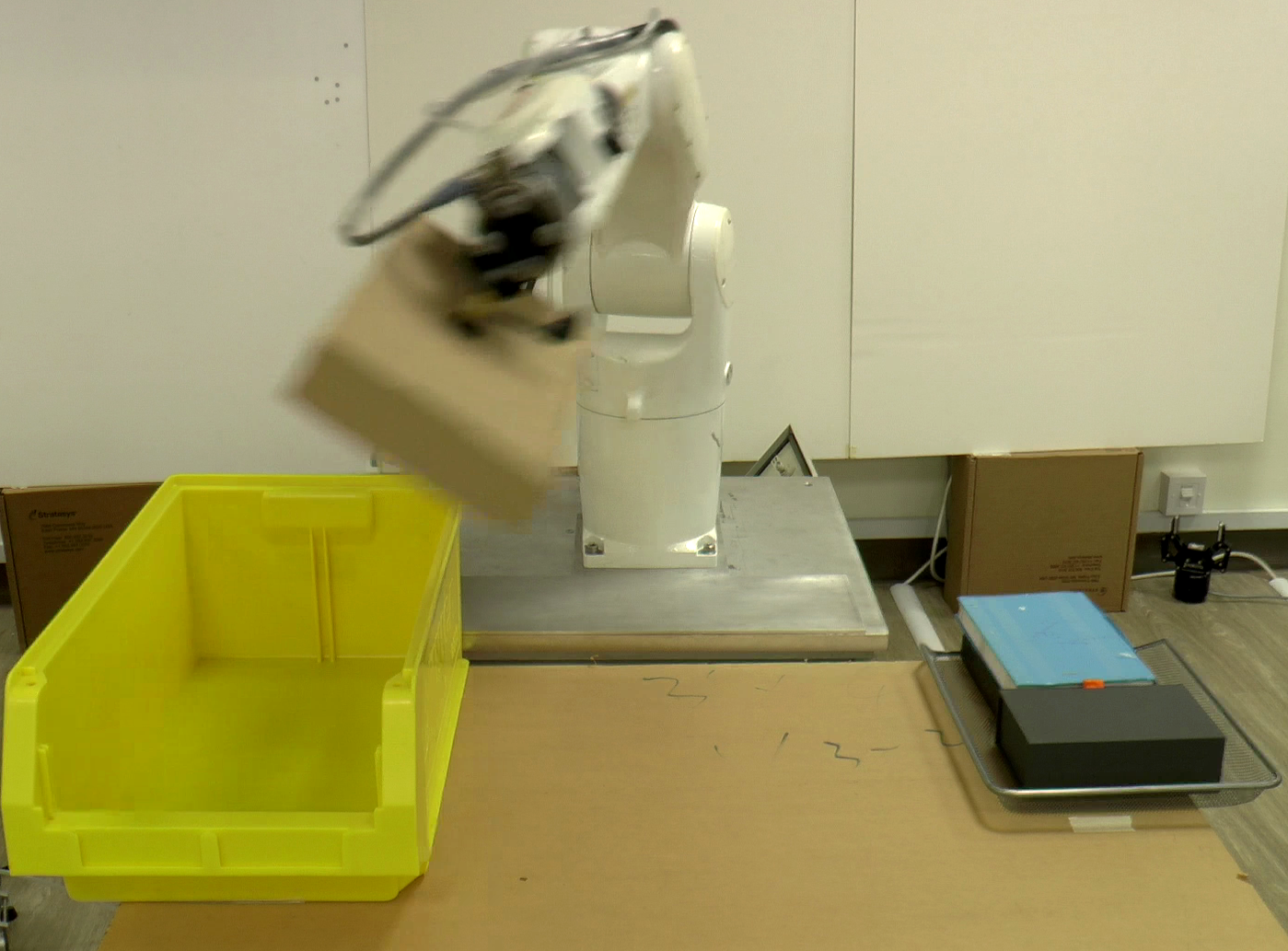}
  \caption{\label{fig:quant} Critically fast pick-and-place of a box
    with a suction cup. The full video of the experiment is available
    at \url{https://youtu.be/b9H-zOYWLbY}.}
\end{figure}

An equally important concern is that the computational cost of
planning should be small, as otherwise it would undermine the benefit
of controlling the robot to move critically fast. For example, suppose
one is capable of computing the true time-optimal trajectories, but
requires a few seconds per trajectory. In this case, it might be more
reasonable to use a sub-optimal approach that has cheaper
computational cost, such as limiting the robot's joint velocity and
acceleration. In the same way, the planning procedure needs to be
reliable: it has to detect infeasible instances correctly, and is
robust against numerical errors.

In this paper, we propose an approach for planning critically fast
movements for robotic pick-and-place with suction cups. This approach
is computationally efficient and robust, and is verified
experimentally to be capable of producing near time-optimal movements.
This performance is achieved by means of three main technical
contributions:
\begin{itemize}
\item A model for suction cup contacts
  (Section~\ref{sec:model-contact-forces});
\item A procedure to compute and approximate the contact stability
  constraint based on that model
  (Section~\ref{sec:strat-simpl-result});
\item A procedure to parameterize, in a time-optimal manner, arbitrary
  geometric paths under the identified contact stability constraint
  (Section~\ref{sec:time-param-with}).
\end{itemize}

The full pipeline is available as
open-source\footnote{\url{https://github.com/hungpham2511/rapid-transport}}. Experimental
results are reported and discussed in Section~\ref{sec:experiment}. A
discussion of related works is postponed to
Section~\ref{sec:related-works}.

\section{Grasp Stability for suction cups}
\label{sec:contact-stability}

\subsection{Background 1: Linearized friction cone}
\label{sec:prel-1:-line}

% [1] https://disopt.epfl.ch/files/content/sites/disopt/files/shared/Strong%20relaxations%2016/7.pdf

In suction cup grasping, frictional forces exist between the suction
cup's pad and the object; these forces are fundamental for maintaining
a stable grasp. These frictional forces are modelled using the Colomb
friction model. Let $\mathbf f = (f_{x}, f_{y}, f_{z})$ denote a
friction force vector, $\mu$ denote the coefficient of friction and
let the Z-axis be the normal contact direction, the Colomb friction
model states
\begin{equation}
  \label{eq:3}
  \|(f_{x}, f_{y})\|_{2} \leq \mu f_{z}.
\end{equation}
The set of feasible friction forces, or equivalently the feasible set
of inequality~\eqref{eq:3}, is a Second-Order cone\footnote{Also
  known as a Lorentz cone or an ice-cream
  cone.}~\cite{Boyd:2004:CO:993483} and hence, can be approximated by a
polyhedral cone with arbitrary precision~\cite{Ben-Tal2001}. For
instance, an approximation with 4 linear inequalities is given as
follows:
\begin{equation}
  \label{eq:19}
  \begin{bmatrix}
    -1 & -1 & -\mu  \\ 
    -1 &  1 & -\mu  \\
     1 &  1 & -\mu  \\
     1 & -1 & -\mu  \\
   \end{bmatrix}
   \begin{bmatrix}
     f_{x} \\ f_{y} \\ f_{z}
   \end{bmatrix} \leq \mathbf 0_{4}.
\end{equation}
This approximation is known as the \emph{linearized friction
  cone}. This is also the approximation we use in this paper.

\subsection{Background 2: Polyhedral computations}
\label{sec:prel-2:-polyt}

A polytope can be defined as the feasible set of a system of linear
equalities and inequalities:
\begin{equation*}
  \mathbf A \mathbf x \leq \mathbf b, \; \mathbf C \mathbf x = \mathbf d,
\end{equation*}
where $\mathbf A, \mathbf b, \mathbf C, \mathbf d$ are matrices of
suitable dimensions. This representation is known as the
H-representation, where H stands for halfspace.  Alternatively, a
polytope can also be defined as the Minkowski sum\footnote{The
  Minkowski sum of two sets $P, Q$ is defined as
  \[\{x + y \mid x \in P, y \in Q\}\]} of the convex hull of a finite
number of points and the conic hull of a finite number of rays:
\begin{equation*}
  \left\{\sum_{i}\theta_{i} \mathbf u_{i} \mid \sum_{i}\theta_{i} = 1; \forall i, \theta_{i} \geq 0 \right\}
  + 
  \left\{\sum_{i}\theta_{i} \mathbf v_{i} \mid \forall i, \theta_{i} \geq 0 \right\}.
\end{equation*}
The vertices $\mathbf u_{i}$ (resp. rays $\mathbf v_{i}$) are referred
to as the \emph{generating} vertices (resp. rays).  This
representation is known as the V-representation, where V stands for
vertex.

A seminal result in the theory of polyhedral computation, discovered
by Minkowski and Weyl, is that any polytope has both a
H-representation and a V-representation~\cite{fukuda1996double}.  The
problem of computing the V-representation given the H-representation
is \emph{vertex enumeration}; and the dual problem is \emph{facet
  enumeration}. Even though both problems are known to be NP-hard,
algorithms, such as the Double Description
method~\cite{fukuda1996double} or the Reversed Search
algorithm~\cite{Avis1996}, have been developed to solve them in
reasonable times. Both algorithms have publicly available
implementations.

\subsection{Exact suction cup grasp stability constraint}
\label{sec:model-contact-forces}

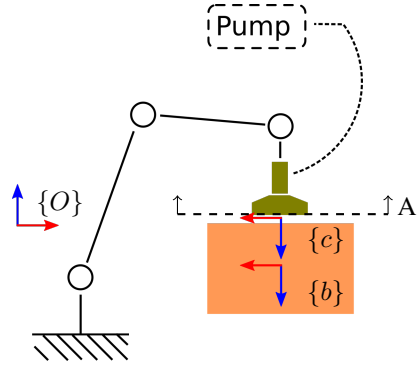
\begin{figure}
  \centering
  \begin{tikzpicture}
    % \draw [step=1] (0, 0) grid (5, 5);

    \node[anchor=south west, inner sep=0] (image) at (0, 0) {\includegraphics[]{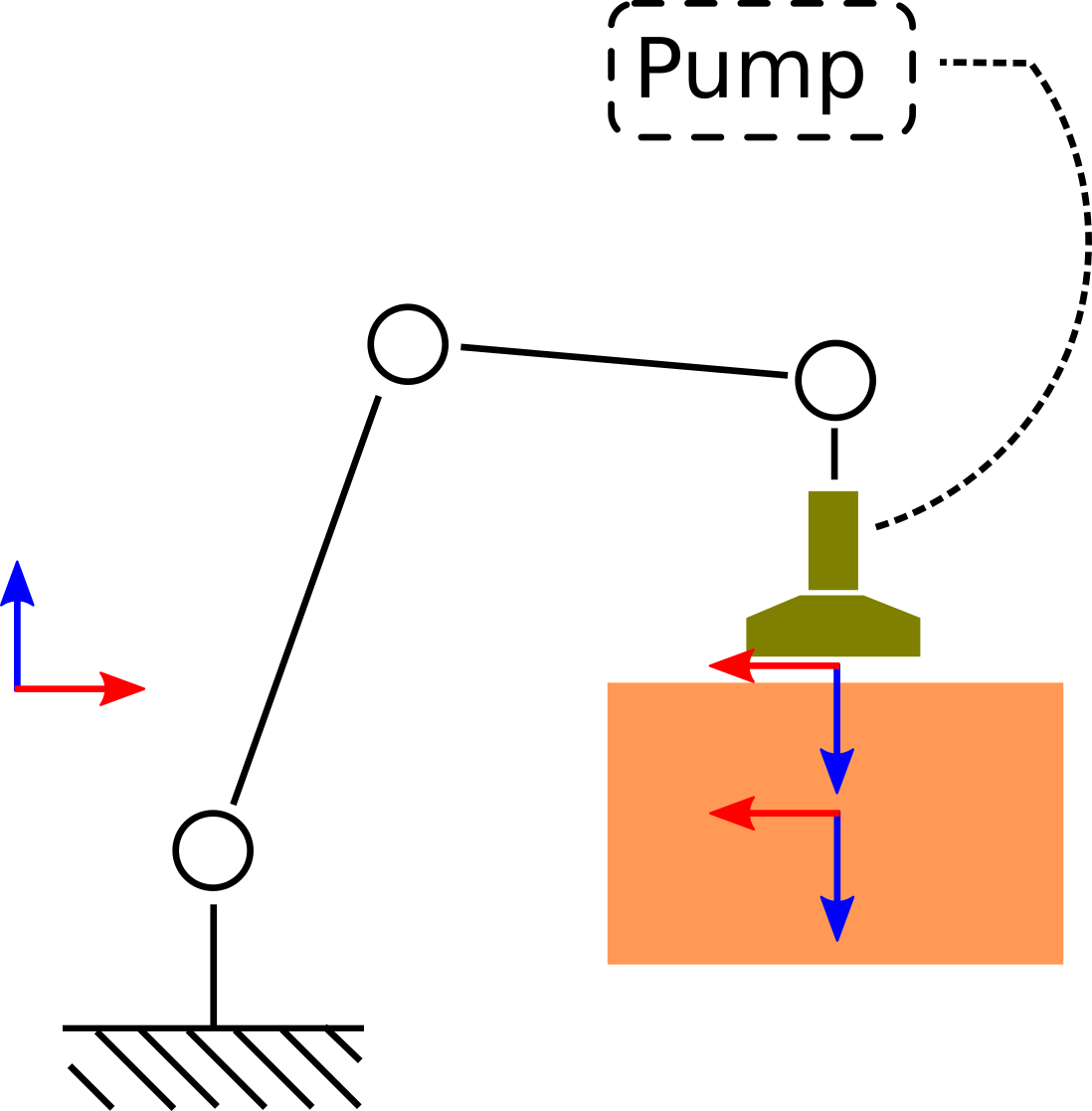}};

    \node[anchor=south west] at (0.2, 1.8) {$\{O\}$};
    \node[anchor=south west] at (3.8, 1.3) {$\{c\}$};
    \node[anchor=south west] at (3.8, 0.6) {$\{b\}$};
    
    \draw[dashed, thick] (2.2, 1.95)--(5.0, 1.95);
    \draw[->] (2.2, 1.95) -- (2.2, 2.2);
    \draw[->] (5.0, 1.95) -- (5.0, 2.2);
    \node[anchor=west] at (5.0, 2.05) {A};

  \end{tikzpicture}
  \caption{\label{fig:sys-visualization} Diagram of a robotic system
    for pick-and-place. Here $\{O\}, \{b\}, \{c\}$ denote respectively
    the global frame, a frame attached to the suction cup and a frame
    attached to the object. A is a section view shown in
    Fig.~\ref{fig:simp-model}.}
\end{figure}

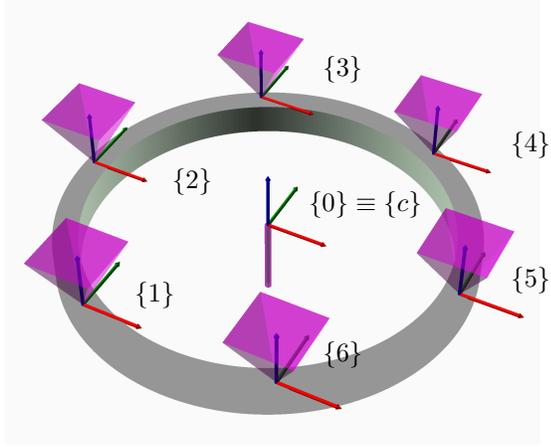
\begin{figure}
  \centering
  \begin{tikzpicture}
    \node[anchor=south west, inner sep=0] (image) at (0, 0) {
      \includegraphics[trim={3cm 5cm 3cm 12cm},clip,width=0.4\textwidth]
      {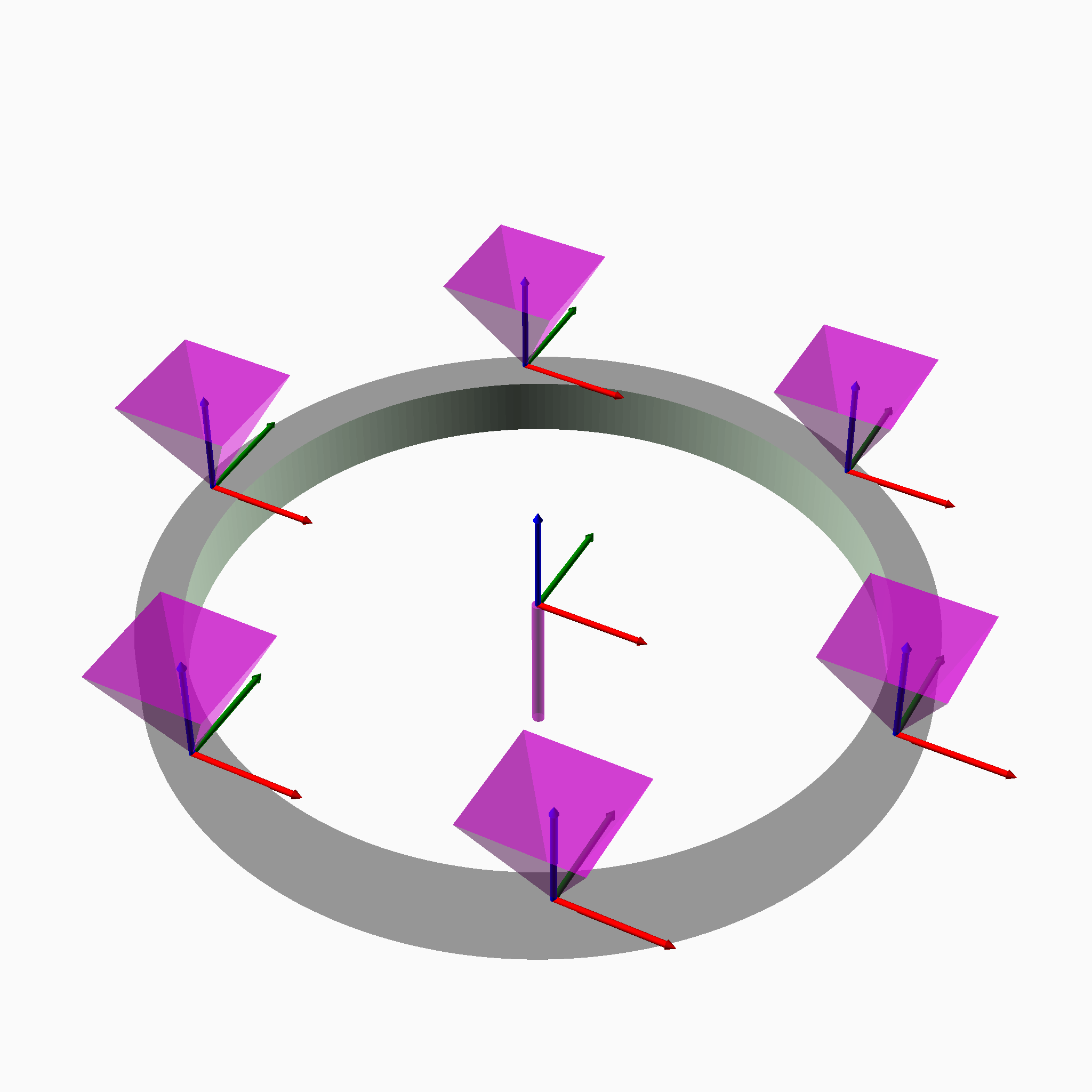}};

    \node[] at (2, 2)     {$\{1\}$};
    \node[] at (2.5, 3.5) {$\{2\}$};
    \node[] at (4.5,   5) {$\{3\}$};
    \node[] at (7. , 4.0) {$\{4\}$};
    \node[] at (7. , 2.2) {$\{5\}$};
    \node[] at (4.5, 1.2) {$\{6\}$};
    \node[] at (4.8, 3.2) {$\{0\}\equiv\{c\}$};
    
    % \draw [step=1] (0, 0) grid (6, 6);
  \end{tikzpicture}
  \caption{\label{fig:simp-model} Cross-sectional view (cross section
    A in Fig.~\ref{fig:sys-visualization}) of the contact between
    the cup and the object. The physical contact area is modelled by
    $m$ point forces following the linearized Colomb friction model
    and 1 point force capturing the suction force due to negative
    pressure. A local frame is defined at each point contact. }
\end{figure}

We model the suction cup-object contact as consisting of multiple
point-wise forces: one suction force and $m$ contact point
forces\,\footnote{From~\cite{caron2015stability}, adopting the
  point-force formulation does not incur any loss in generality as
  compared to the surface-force formulation} (See
Fig.~\ref{fig:simp-model}). Let $\mathbf f^{(0)}$ be the suction force
in frame $\{0\}$, which is attached to the centroid of the suction
cup, one has:
\begin{equation}
  \label{eq:9}
  \mathbf f^{(0)}_{0} = [0, 0, PA]^{\top},
\end{equation}
where $P$ is the negative pressure in $\SI{}{Pa}$ and $A$ is the
suction area in $\SI{}{m^{2}}$.  Let $\mathbf f^{(i)}$ denote the
$i$-th contact force exerted on the object in the contact local frame.
We assume that $\mathbf f^{(i)}$ follows the linearized Colomb
friction model~\eqref{eq:19}.  The suction and contact forces with the
corresponding local frames $(\{i\}, i=0,\dots,m)$ are depicted in
Fig.~\ref{fig:simp-model}.

The net contact wrench exerted on the object by the cup depends
linearly on these point forces.  Indeed, let $\{c\}$ denote the frame
attached rigidly to the suction cup.  By transforming all individual
forces to frame $\{c\}$ followed by summation, one obtains the net
wrench $\mathbf w_{c}$:
\begin{equation}
  \label{eq:8}
  \mathbf w_{c} = \sum_{i=0}^{m}
  \mathbf G_{ci} \mathbf f_{i}^{(i)},
\end{equation}
where the matrix $\mathbf G_{ci}$ is a \emph{fixed} $6\times3$ matrix
that transforms a force in frame $\{i\}$ to a wrench in frame $\{c\}$.
We can compute $\mathbf G_{ci}$ from the position vector
$\mathbf p_{ci}$ (position of the origin of frame $\{i\}$ in frame
$\{c\}$) and rotational matrix $\mathbf R_{ci}$ as below
\begin{equation*}
  \mathbf G_{ci} =
  \begin{bmatrix}
    [\mathbf p_{ci} \times] \mathbf R_{ci}\\
    \mathbf R_{ci}
  \end{bmatrix}.
\end{equation*}

We make the following key assumption concerning contact stability for
suction cup.  This condition is often used in humanoid
locomotion~\cite{Caron2017} and is known as the \emph{weak contact
  stability} condition~\cite{pang2000stability}.
\begin{assum}[Contact stability]
  \label{assum:1}
  A contact wrench $\mathbf w_{c}$ is physically realizable, and hence
  allows stable suction-cup object grasp stability, if there exists a
  set of individual contact forces
  $\mathbf f^{(0)},\dots,\mathbf f^{(m)}$ that satisfies
  Eq.~\eqref{eq:19} and~\eqref{eq:9}.
\end{assum}

An important result that follows from Assump.~\ref{assum:1} is that
there exists $\mathbf F_{c}$ and $\mathbf g_{c}$ such that if the
contact wrench satisfies
\begin{equation}
  \label{eq:20}
  \mathbf F_{c} \mathbf w_{c} \leq \mathbf g_{c},
\end{equation}
contact stability is achieved.  We prove this result by the below
constructive procedure, which shows how one can compute
$\mathbf F_{c}$ and $\mathbf g_{c}$ directly:
\begin{enumerate}
\item let $\hat {\mathcal F}$ be the set of feasible values of
  $\hat {\mathbf f} :=\{\mathbf f^{(0)},\dots,\mathbf f^{(m)}\}$;
  form the H-representation of $\hat{\mathcal F}$;
\item compute the corresponding V-representation, which is a set of
  generating vertices\footnote{There can not be generating rays since
    $\hat{\mathcal F}$ is bounded.}
  $(\hat{\mathbf f}_{1},\dots,\hat{\mathbf f}_{l})$;
\item use Eq.~\eqref{eq:8} to transform
  $(\hat{\mathbf f}_{1},\dots,\hat{\mathbf f}_{l})$ to
  $(\hat{\mathbf w_{1}},\dots,\hat{\mathbf w_{l}})$; $\mathcal W_{c}$
  is the convex hull of 
  $(\hat{\mathbf w_{1}},\dots,\hat{\mathbf w_{l}})$;
  
\item compute the H-representation of $\mathcal W_{c}$, which are the
  coefficient matrices $\mathbf F_{c}, \mathbf g_{c}$ in
  Eq.~\eqref{eq:20}.
\end{enumerate}
Refer to Section~\ref{sec:prel-2:-polyt} for a discussion on
transforming between the two representations (H and V) of a polytope.

\subsection{Approximate suction cup grasp stability constraint}
\label{sec:strat-simpl-result}

\begin{figure}
  \centering
  \begin{tikzpicture}
    \node[anchor=south west, inner sep=0] (image) at (0, 0)
    {\includegraphics[]{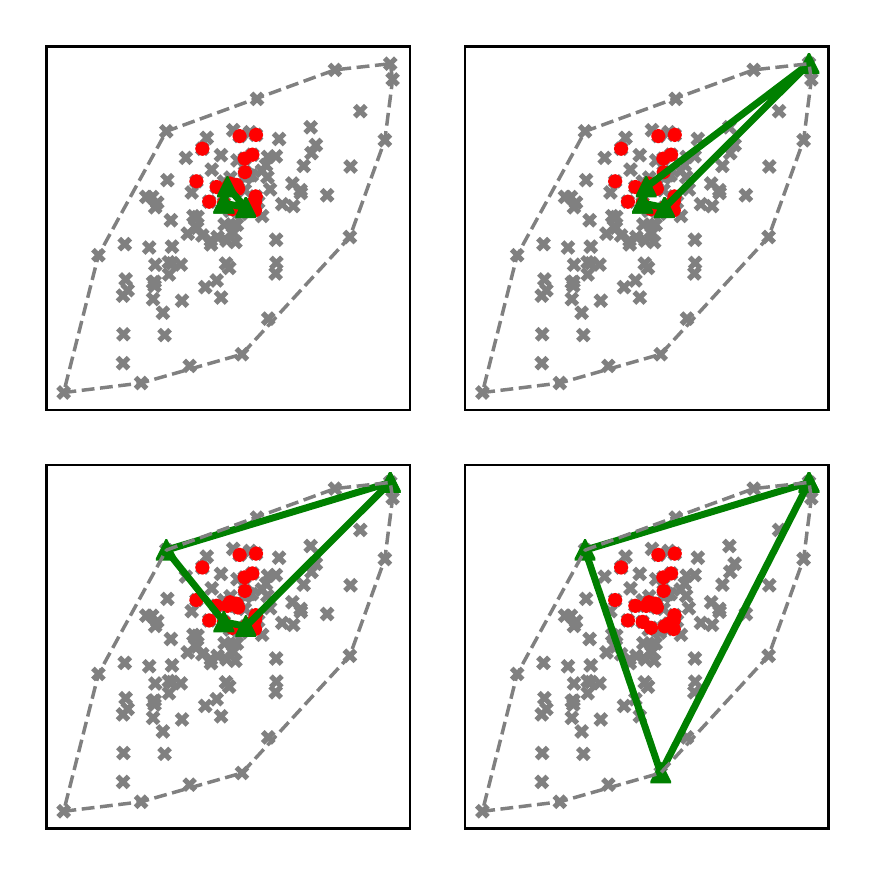}};
    \node[anchor=north west] at (0.5, 8.4) {\bf iter 0};
    \node[anchor=north west] at (4.8, 8.4) {\bf iter 1};
    \node[anchor=north west] at (0.5, 4.2) {\bf iter 2};
    \node[anchor=north west] at (4.8, 4.2) {\bf iter 3};
  \end{tikzpicture}
  
  \caption{Application of the procedure proposed (to approximate grasp
    stability constraint) on a planar data set.  Vertices
    $\hat{ \mathbf w_{i}}$, guiding samples are shown as gray crosses
    and red dots respectively. The proposed procedure finds 3 vertices
    (iter 3) whose convex hull covers the guiding samples
    completely. The dashed grey polygon, representing the exact
    constraint, has four times as many edges.}
  \label{fig:square}
\end{figure}
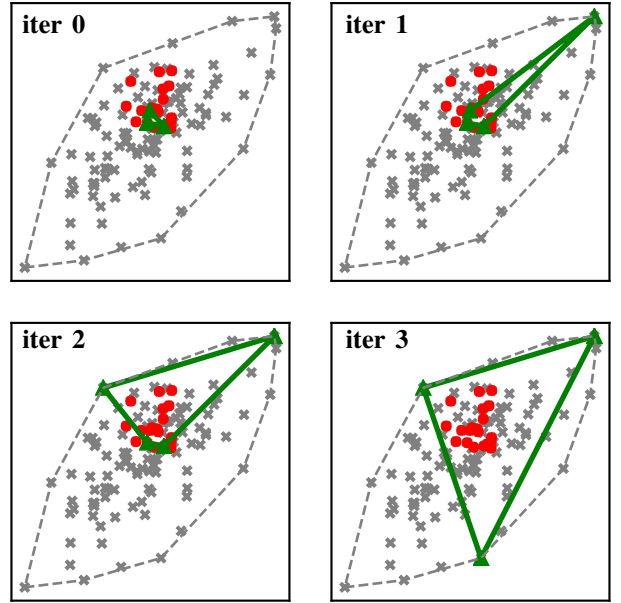

In practice, $\mathbf F_{c}, \mathbf g_{c}$ have many rows, causing
computational difficulties.  We alleviate this issue by approximating
the exact constraint.  The key idea is to represent $\mathcal W_{c}$
as the convex hull $\mathcal H$ of \emph{a set} of points $\mathcal Y$
that (i) belong to $\mathcal W_{c}$ and such that (ii) $\mathcal H$
contains contact wrenches that are \emph{the most likely to be
  realized during execution}. To achieve (ii), we randomly sample
\emph{guiding samples}--contact wrenches that likely occur in
execution. This can be done by either sampling real execution data or
simulation data.

Concretely, the procedure proceeds as follows:
\begin{enumerate}
\item randomly sample $N$ guiding samples; only samples that belongs
  to $\mathcal W_{c}$ are retained; 
\item initialize $\mathcal Y$ as a simplex with 7 randomly chosen
  guiding samples as vertices;
\item compute the convex hull  $\mathcal H$ of $\mathcal Y$;
\item choose the face $h^{*}$ of $\mathcal H$ that contains the most
  guiding samples in its infeasible halfspace;
\item add to $\mathcal Y$ the vertex in $\mathcal W_{c}$ that is in
  the infeasible halfspace of $h^{*}$ and is furthest away from it;
\item repeat from step 3 until the number of vertices in
  $\mathcal Y$ is greater than a specified value.
\end{enumerate}
Remark that in step 2), since the wrench space is 6 dimensional, a
simplex in this space has 7 vertices.  After each iteration, a point
in
$(\hat{\mathbf w_{1}},\dots,\hat{\mathbf w_{l}}) \setminus \mathcal Y$
is added to $\mathcal Y$. In our experiments, after about 50
iteration, the convex hull $\mathcal H$ covers a significant portion
of the guiding samples and has approximately $3000$-$4000$ faces.  In
contrast, $\mathcal W_{c}$ can have up to $150000$ faces.

An application of this procedure on a planar data set is demonstrated
in Fig.~\ref{fig:square}. The green triangle, representing the
approximate grasp constraint, has only 3 edges but covers all guiding
samples.

\section{Planning critically fast movements}
\label{sec:time-param-with}

\subsection{Motion planning pipeline}
\label{sec:moti-plann-pipel}

We propose a motion planning pipeline following the Path-Velocity
Decomposition principle~\cite{kant1986toward}:
\begin{enumerate}
\item Find a collision-free path using standard geometric planners
  (e.g. RRT~\cite{kuffner2000rrt});
\item Time-parameterize the collision-free path to minimize traversal
  time under kinodynamic constraints: joint velocity bounds, joint
  acceleration bounds and suction cup constraints.
\end{enumerate}

% In this approach, a collision-free robot trajectory that
% satisfies given kinodynamic constraints is found in two stages.
% Consider a $n$-dof fully-actuated robotic manipulator whose joint
% positions are denoted by $\mathbf q\in \mathbb R^{n}$, in the first
% stage, a collision-free geometric path
% \begin{equation*}
% \mathbf p(s)_{s\in[0, 1]} \in \mathbb R^{n}
% \end{equation*}
% is planned using a standard geometric path planner (e.g. RRT). In the
% second stage, the geometric path $\mathbf p(s)$ is retimed subject to
% all constraints on the robot system dynamics. That is, we compute the
% time-optimal path-parameterization, which is an increasing scalar
% function of time $s(t)_{t\in [0, T]}\in [0, 1]$, so that the
% trajectory defined by
% \begin{equation*}
%  \mathbf q(t)_{t\in [0, T]}:=\mathbf p(s(t)) 
% \end{equation*}
% satisfy all constraints, which consist of joint velocity, acceleration
% bounds and suction cup constraints and that the duration $T$ is
% minimal. The result is a trajectory that is both collision-free and
% satisfies all given constraints.

Although the decoupling approach does not generally produce optimal
nor even locally-optimal trajectories, it has the distinctive
advantage of being robust and fast. This is due to the highly mature
states of collision-free path planning and path time-parameterization
techniques~\cite{pham2017admissible}.

Regarding the latter, the problem of finding the time-optimal
time-parameterization of a geometric path subject to kinodynamic
constraints is a classical problem in
robotics~\cite{bobrow1985time}. If the constraints under consideration
are of First- or Second-Order (see below and also in~\cite{Pham2017}),
then this problem can be solved extremely efficiently using the
recently-developed Time-Optimal Path Parameterization via Reachability
Analysis (TOPP-RA) algorithm~\cite{Pham2017}. Interested readers
please refer to the paper~\cite{Pham2017} or the open-source
implementation at
\href{https://github.com/hungpham2511/toppra}{https://github.com/hungpham2511/toppra}.

In the next section, we show that the grasp stability constraint is a
Second-Order constraint.

\subsection{Grasp stability constraint as Second-Order constraint}
\label{sec:grasp-stab-constr}

A Second-Order constraint has the following form~\cite{Pham2017}
\begin{equation}
  \label{eq:13}
  \mathbf A(\mathbf q) \ddot {\mathbf q} + \dot{\mathbf q}^{\top}\mathbf B(\mathbf q) \dot{\mathbf q}
  + \mathbf f(\mathbf q) \in \mathscr C(\mathbf q),
\end{equation}
\begin{itemize}
\item $\mathbf q$ is the vector of joint values of the robot;
\item $\mathbf{A, B, f}$ are continuous mappings from $\mathbb{R}^n$
  to $\mathbb{R}^{m\times n},\mathbb{R}^{n\times m\times n}$ and
  $\mathbb{R}^{m}$ respectively;
\item $\cC(\vec q)$ is a convex polytope in $\mathbb{R}^{m}$.
\end{itemize}
To establish that any grasp stability constraint can be reformulated
as a Second-Order constraint, recall first the following relationships
between the robot's joint position and the object's
motion~\cite{Hourtash2005a}:
\begin{align}
  \label{eq:11}
    \bm \omega_{b} &= \mathbf J_{\rm rot}(\mathbf q) \dot {\mathbf q}, & {\bm \alpha}_{b} &= \mathbf J_{\rm rot}(\mathbf q) \ddot {\mathbf q} + \dot{\mathbf q}^{\top}\mathbf H_{\rm rot}(\mathbf q) \dot{\mathbf q},\\
  \label{eq:5}
    \mathbf v_{b} &= \mathbf J_{\rm trans}(\mathbf q) \dot {\mathbf q}, & \mathbf a_{b} &= \mathbf J_{\rm trans}(\mathbf q) \ddot {\mathbf q} + \dot{\mathbf q}^{\top}\mathbf H_{\rm trans}(\mathbf q) \dot{\mathbf q},
\end{align}
where $\mathbf J_{\Box}, \mathbf H_{\Box}$ are the translational and
rotational Jacobians and Hessians, $\mathbf v_{b}, \bm \omega_{b}$
denote the object's translational and rotational velocities and
$\mathbf a_{b}, \bm \alpha_{b}$ denote the translational and
rotational accelerations; all are in the object's body frame $\{b\}$.
Next, combining Eq.~\eqref{eq:11} and~\eqref{eq:5} with the
Newton and Newton-Euler equations in the body frame, which are
\begin{equation}
  \label{eq:6}
  \mathbf w_{b} + \begin{bmatrix} 0_{3} \\ \mathbf g_{b} m \end{bmatrix}
  =
  \begin{bmatrix}
  \mathbf I_{b} \bm \alpha_{b} + \bm \omega_{b} \times \mathbf I_{b} \bm \omega_{b} \\
  m \mathbf a_{b},
  \end{bmatrix},
\end{equation}
where $m$ is the mass of the object, $\mathbf w_{b}$ is the
interaction wrench, $\mathbf g_{b}$ is gravitational acceleration,
$\mathbf I_{b}$ is the inertia matrix of the object both in the
object's body frame $\{b\}$.  Rearranging the terms of Eq.~\ref{eq:6}
and substituting in Eq.~\eqref{eq:11} and~\eqref{eq:5}, one can show
that $\mathbf w_{b}$ has the form
\begin{equation*}
  \mathbf w_{b} =
  \Theta_{1}(\mathbf q)\ddot {\mathbf q} + \dot{\mathbf {q}}^{\top}
  \Theta_{2}(\mathbf q) \dot{\mathbf q} + \Theta_{3}(\mathbf q),
\end{equation*}
where $\Theta_{1},\Theta_{2},\Theta_{3}$ are tensors depending on the
geometry of the robot and the inertial properties of the object.

Next, let $\mathbf G_{cb}$ be the \emph{constant} matrix that
transforms a wrench from the object's body frame $\{b\}$ to the
suction cup's frame $\{c\}$. Substituting to Eq.~\eqref{eq:20},
grasp stability constraint can then be written as
\begin{equation}
  \label{eq:1}
  \mathbf F_{c} \mathbf G_{cb}\left\{\Theta_{1}(\mathbf q)\ddot {\mathbf q} + \dot{\mathbf {q}}^{\top}
  \Theta_{2}(\mathbf q) \dot{\mathbf q} + \Theta_{3}(\mathbf q)\right\} \leq \mathbf g_{c}.
\end{equation}
We can see that Eq.~\eqref{eq:1} is a Second-Order constraint
according to the definition given earlier. Indeed,
$\Theta_{1}, \Theta_{2}, \Theta_{3}$ correspond to matrices
$\mathbf {A, B, f}$ respectively and the fixed convex polytope
\begin{equation*}
  \{\mathbf w_{b} \mid \mathbf F_{c} \mathbf G_{cb} \mathbf w_{b} \leq \mathbf g_{c}\}.
\end{equation*}
corresponds to $\mathscr C(\mathbf q)$.

\section{Experiments}
\label{sec:experiment}

% \begin{figure}
%   \centering
%   \begin{tikzpicture}
%     \draw (0, 0) rectangle (5, 6);
%   \end{tikzpicture}
%   \caption{\label{fig:exp-setup} Experimental setup. A: Robotic
%     arm. B: Suction cup.  C: Vacuum pump.  }
% \end{figure}

Two aspects of the proposed approach were experimentally
investigated. First, we evaluate the \emph{quality} of the
trajectories for transporting object by looking at the rate of
successful transport and the trajectories' durations in 20 randomly
generated instances (\ref{sec:experiment}-A). Second, we report the
actual computational cost of the proposed pipeline in a realistic
pick-and-place scenario (\ref{sec:experiment}-B).

\subsection{Experimental setup}
\label{sec:setup}

The same equipment was employed throughout the experiments. These
include a position-controlled industrial robot Denso VS-060, equipped
with a suction cup connected to a vacuum pump. The robot is controlled
at $\SI{125}{Hz}$.  The suction cup has a radius of $\SI{12.5}{mm}$
and the vacuum pump generates a negative pressure of approximately
$\SI{30}{kPa}$.  Objects considered in the experiments have varying
weights ranging from $\SI{0.2}{kg}$ to $\SI{0.6}{kg}$; all objects
have known weights and moment of inertia. All computations were done
on a single core of a laptop running Ubuntu 16.04 at
$\SI{3.800}{GHz}$.

Collision-free paths between two robot configurations were all
computed using \texttt{OpenRAVE}'s implementation~\cite{Diankov2008}
of \texttt{biRRT} algorithm~\cite{kuffner2000rrt}.

A single Suction Cup stability Constraint (SSC)
$\mathbf F_{c} \mathbf w \leq \mathbf g_{c}$ was identified and
approximated \emph{offline}, following the procedure given in
Section~\ref{sec:contact-stability}. The following parameters were
used: $m=6$ points to approximate the suction cup-object contact area;
coefficient of friction $\mu=0.3$, identified using a Force-Torque
sensor; suction cup radius $\SI{12.5}{mm}$; maximum $60$ vertices in
the V-representation of the constraint
(cf. Sec.~\ref{sec:strat-simpl-result}). The approximated coefficient
matrices have \emph{$3099$ rows}, while the exact coefficient matrices
have $130000$ rows. All TOPP instances were discretized with $100$
gridpoints, using the first-order interpolation
scheme~\cite{Pham2017}. The code is written mostly in Python and
Cython. TOPP-RA is configured to run a custom LP solver based on
Seidel's algorithm~\cite{seidel1991small}.

\subsection{Trajectory quality}
\label{sec:robustness}

We first looked at the likelihood of the objects falling, slipping or
twisting when executing the trajectories found by our pipeline. As a
baseline for comparison, we also considered an alternative strategy,
where one computes time-optimal time-parameterizations subject only to
the robot's kinematic constraints, without taking into account the
suction cup constraint.

Twenty geometric paths were randomly generated and tested. To mimic
actual pick-and-place settings, we adopted the following procedure:

\begin{enumerate}
\item sample randomly the object's starting and goal poses, each in
  a dedicated region of the workspace;
\item use \texttt{OpenRAVE}'s inverse kinematics to find the
  corresponding robot's starting and goal configurations;
\item find a path between the starting and goal configurations using
  \texttt{OpenRAVE}.  
\end{enumerate}

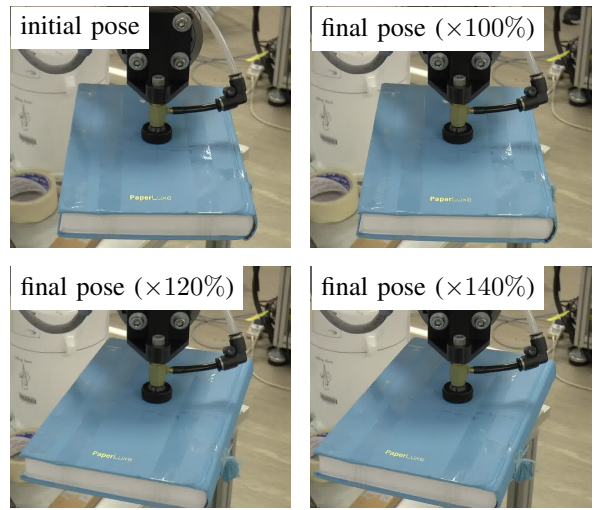
\begin{figure}
  \centering
  \begin{tikzpicture}
    \node[anchor=south west,inner sep=0] (image) at (0, 0){
      \includegraphics[trim={16cm 5cm 19cm 5cm}, clip, width=0.21\textwidth] {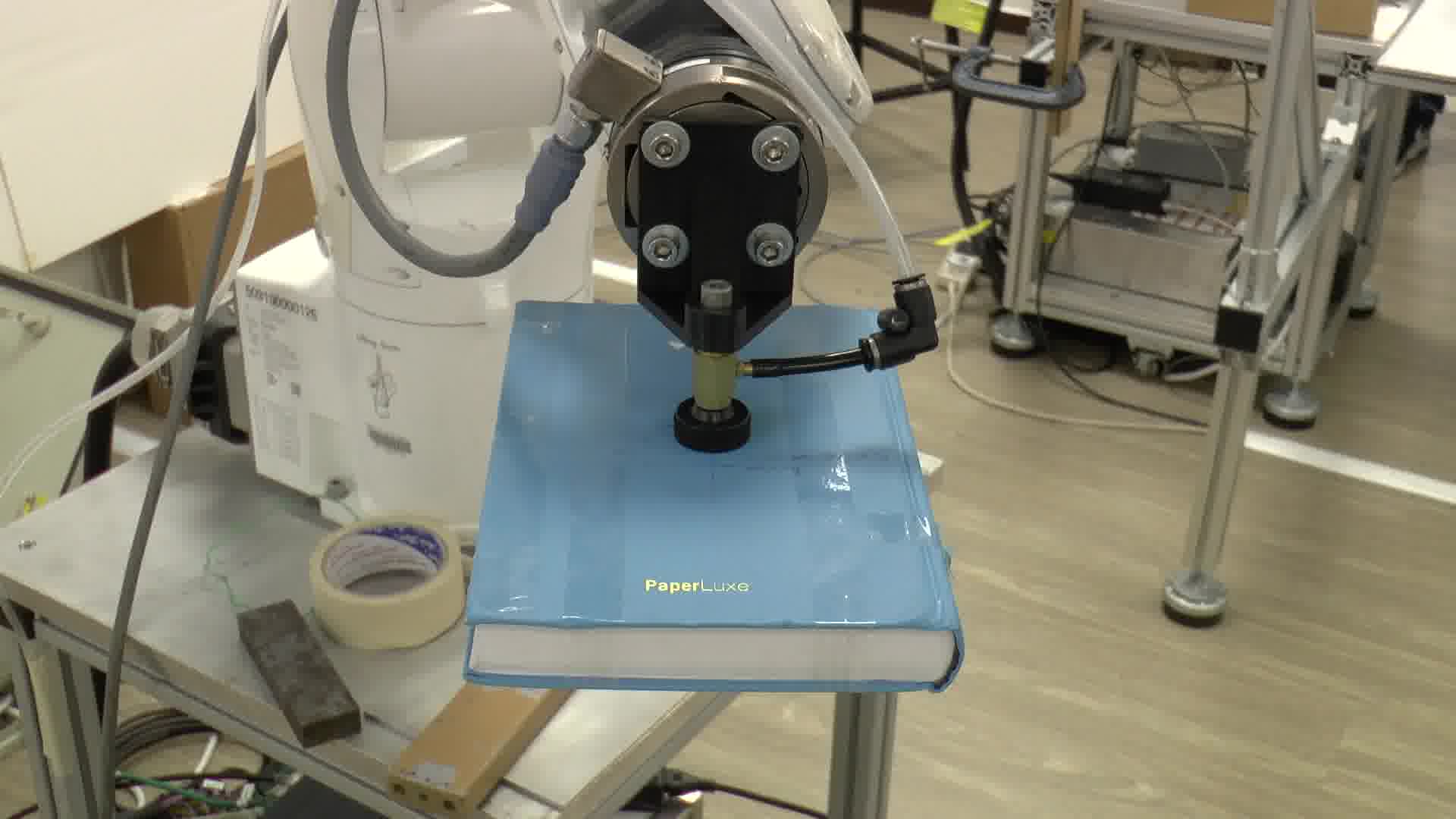}};
    \node[anchor=south west,inner sep=0] (image) at (4, 0){
      \includegraphics[trim={16cm 5cm 19cm 5cm}, clip, width=0.21\textwidth] {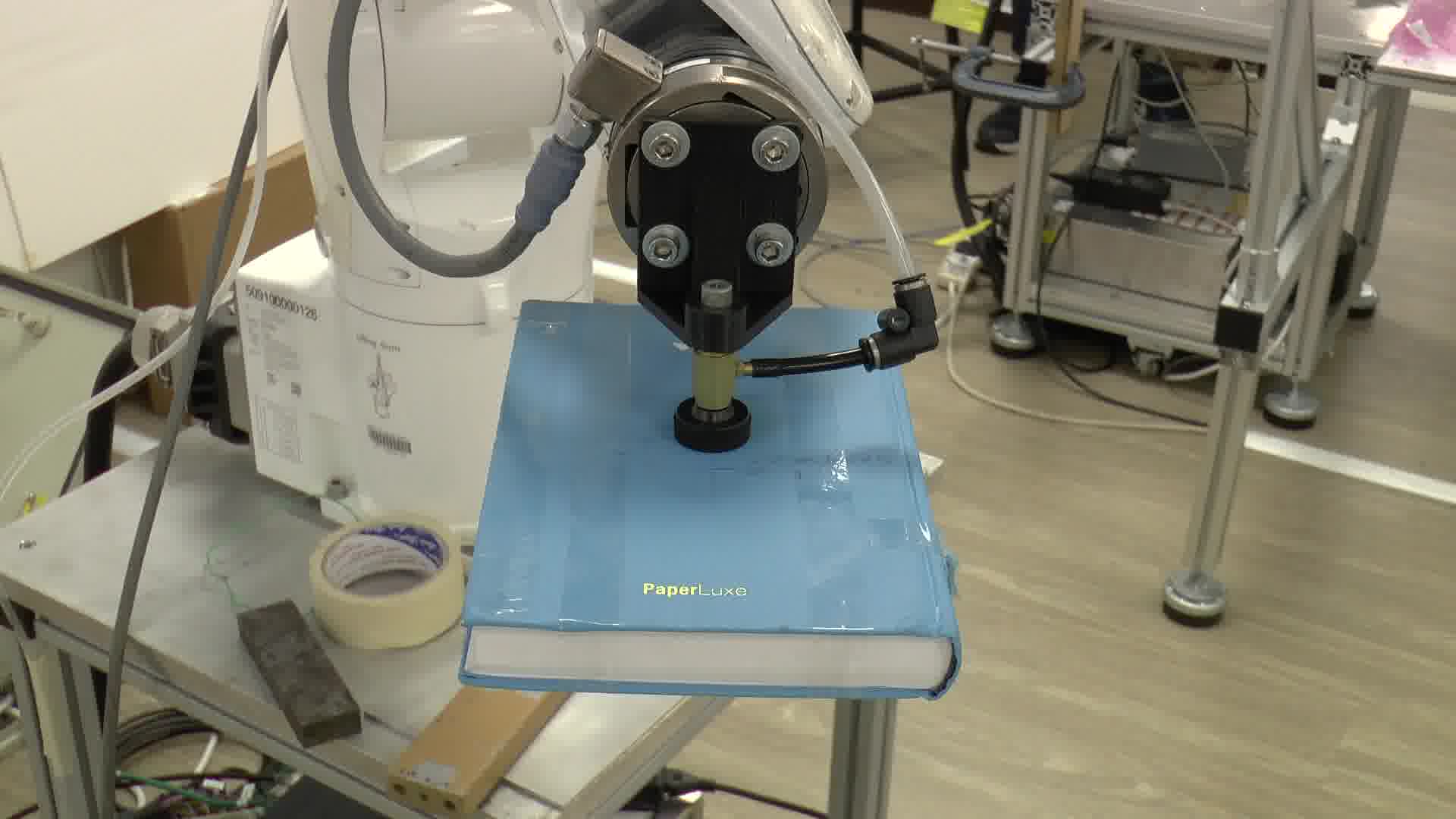}};
    \node[anchor=south west,inner sep=0] (image) at (4, -3.45){
      \includegraphics[trim={16cm 5cm 19cm 5cm}, clip, width=0.21\textwidth] {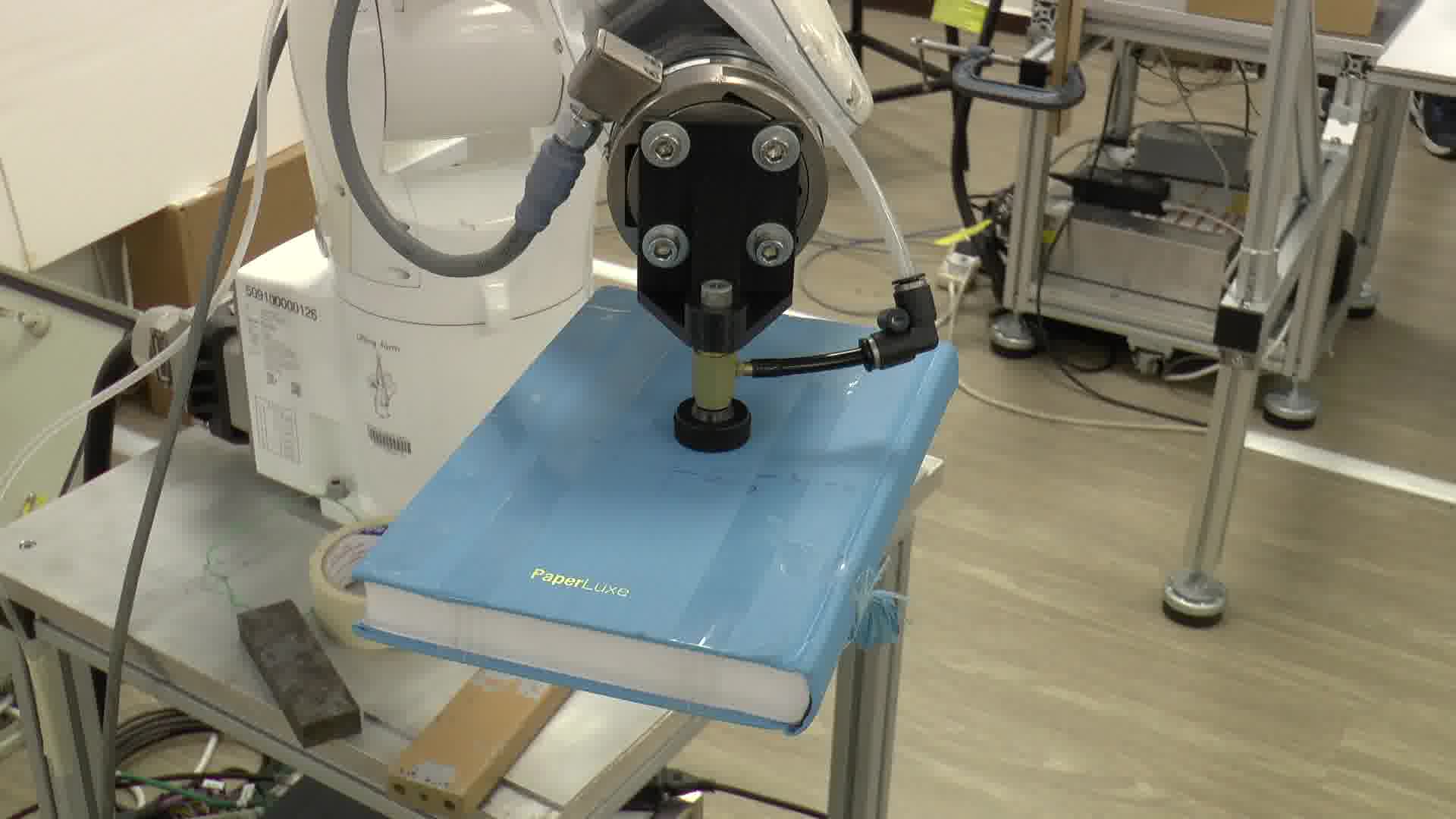}};
    \node[anchor=south west,inner sep=0] (image) at (0, -3.45){
      \includegraphics[trim={16cm 5cm 19cm 5cm}, clip, width=0.21\textwidth] {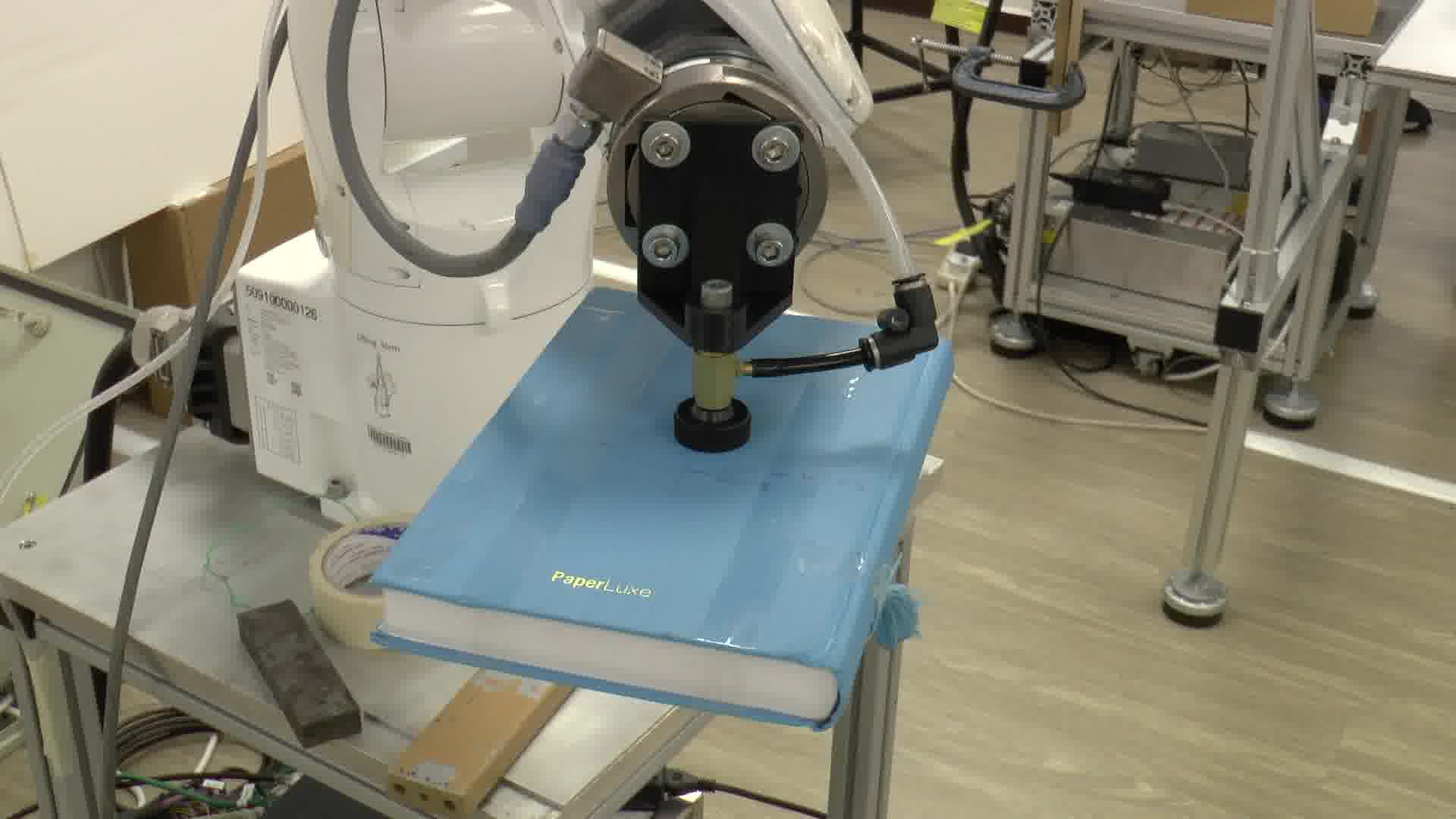}};
    
    \node[anchor=north west, fill=white] at (0.02, 3.21) {initial pose};
    \node[anchor=north west, fill=white] at (4.02, 3.21) {final pose ($\times 100\%$)};
    \node[anchor=north west, fill=white] at (4.02, -0.23) {final pose ($\times 140\%$)};
    \node[anchor=north west, fill=white] at (0.02, -0.23) {final pose ($\times 120\%$)};
  \end{tikzpicture}
  \caption{\label{fig:suboptimality} Executing a retimed trajectory at
    $120\%$ and $140\%$ speed led to visible twisting and slipping.}
\end{figure}

In all trials, a rectangle notebook (See Fig.~\ref{fig:suboptimality})
with weight $\SI{0.551}{kg}$ and moment of inertia
$\mathrm{diag}(9.28, 21.10, 29.80)\times\SI{e-4}{kg m^{2}}$ was
transported with the suction cup. The contact point is $\SI{12.5}{mm}$
above the notebook's center of mass.

It was found that, by accounting for SCC, the object could be
transported without slipping or falling for all 20 paths
(Table~\ref{tab:retime-time}). Furthermore, these trajectories were
not significantly slower comparing to the robot's hardware speed
limits.  On the other hand, only 4 out of 20 trajectories retimed
without SSC could be executed successfully. We also observed that 3 in
these 4 successful trajectories violate SSC. This could be because the
linearized frictional constraint~\eqref{eq:19} or the approximation
procedure in Section~\ref{sec:strat-simpl-result} are conservative.

\begin{table}[htp]
  \caption[]{\label{tab:retime-time} Trajectory duration (in seconds)
    and success rate of 20 randomly generated
    trajectories retimed without and with suction cup constraints
    (SCC).}
  \centering
  \begin{tabular}{lrrr}
    \toprule
    \midrule
    \# & without SCC & with SCC   & time ext. (\%) \\
    \midrule
    0  & \fa{0.432}  & \su{0.672} &  35.71         \\
    1  & \fa{0.568}  & \su{0.832} &  31.73         \\
    2  & \fa{0.944}  & \su{1.008} &  6.35          \\
    3  & \fa{0.504}  & \su{0.688} &  26.74         \\
    4  & \fa{0.728}  & \su{0.952} &  23.53         \\
    5  & \fa{0.504}  & \su{0.672} &  25.00         \\
    6  & \su{1.064}  & \su{1.064} &  \bsu{0.00}          \\
    7  & \fa{0.520}  & \su{0.688} &  24.42         \\
    8  & \fa{1.136}  & \su{1.344} &  15.48         \\
    9  & \fa{0.496}  & \su{0.744} &  33.33         \\
    10 & \su{0.632}  & \su{0.728} &  \bsu{13.19}         \\
    11 & \fa{0.520}  & \su{0.648} &  19.75         \\
    12 & \fa{0.440}  & \su{0.728} &  39.56         \\
    13 & \su{0.824}  & \su{0.896} &  \bsu{8.04}          \\
    14 & \fa{1.104}  & \su{1.392} &  20.69         \\
    15 & \fa{0.528}  & \su{0.664} &  20.48         \\
    16 & \fa{0.520}  & \su{0.736} &  29.35         \\
    17 & \fa{1.344}  & \su{1.848} &  27.27         \\
    18 & \fa{0.472}  & \su{0.760} &  37.89         \\
    19 & \su{0.512}  & \su{0.600} &  \bsu{14.67}         \\
    \bottomrule
  \end{tabular} \\
  \vspace{0.2cm}
\end{table}

% This data also suggests that trajectories generated with the proposed
% approach are close to being \emph{physically} optimal, despite
% imperfections due to modelling and approximation errors. Consider
% paths number 6, 10, 13 and 19 in Table~\ref{tab:retime-time}, these
% paths can be executed successfully even without considering grasp
% stability constraints. This implies the corresponding retimed
% trajectories are the true optimal solutions, which have duration
% $\SI{1.064}{sec}, \SI{0.632}{sec}, \SI{0.824}{secs}$ and
% $\SI{0.512}{sec}$ respectively. For these cases, the proposed approach
% produces retimed trajectories with duration that are at most $14.6\%$
% from the optimal values. It follows that the degree of sub-optimality
% is less than $14.6\%$.

A second experiment was performed to examine the degree of
sub-optimality. We took paths number 12 and 18, parameterized both
subject to grasp stability constraint and executed them at three
levels of speed: $100\%, 120\%, 140\%$. We then inspected the pose of
the object relative to the suction cup after each
execution. Significant slipping of nearly $\SI{1}{cm}$ and
$\SI{10}{deg}$ were found at the trials executed at $120\%$ and
$140\%$ speed, see Fig.~\ref{fig:suboptimality} and first part of the
experimental video \url{https://youtu.be/b9H-zOYWLbY}.

\subsection{Computational performance}
\label{sec:results-discussions}

\begin{table}[htp]
  \caption{\label{tab:comp-cost} Breakdown of task cycle times (in
    seconds).}
  \centering
  \begin{tabular}{lrrrr}
    \toprule
                  & obj 1  & obj 2  & obj 3 & obj 4 \\
    \midrule
    APPROACH plan & 0.018& 0.053& 0.016& 0.041\\
    REACH plan    & 0.029& 0.034& 0.040& 0.034\\
    MOVE plan    & 0.183& 0.101& 0.147& 0.198\\
    MOVE retime   & 0.111& 0.111& 0.124& 0.137\\
    \arrayrulecolor{black!30} \midrule
    Total plan   & 0.341& 0.299& 0.327& 0.410\\
    \arrayrulecolor{black}   \midrule
    APPROACH exec. & 0.438& 0.747& 0.550& 0.531\\
    REACH exec.    & 0.306& 0.368& 0.415& 0.436\\
    ATTACH delay & 1 & 1 & 1 & 1\\
    MOVE exec.     & 0.841& 0.729& 1.062& 1.371\\
    DETACH delay & 1 & 1 & 1 & 1\\
    \arrayrulecolor{black!30} \midrule
    Total exec. & 3.585& 3.844& 4.027& 4.338\\
   \arrayrulecolor{black}    \bottomrule
    Total (plan + exec.)~\footnotemark{}      & 3.926& 4.143& 4.354& 4.748\\
   \arrayrulecolor{black}    \bottomrule
  \end{tabular}
\end{table}

\footnotetext{Approximation of constraint coefficients was done
  offline.}

To investigate the computational cost of the proposed approach, we
considered a pick-and-place scenario that involves the robot picking
objects using the suction cup from a bin and stacking them at a
distant goal position (Fig.~\ref{fig:quant}). The objective is to
complete the task as fast as possible, planning and execution time
included. The object properties, initial poses and final desired poses
were determined prior to the experiment. The objects weighted between
$\SI{0.204}{kg}$ and $\SI{0.551}{kg}$.

A pick-and-place cycle consisted of several phases. First, the robot
\emph{approaches} a pose that is directly on top of the bin. Then, it
\emph{reaches} for the topmost object, closes the vacuum valve, waits
for $1$ second for the object to \emph{attach} to the suction cup, and
finally \emph{moves} to the given destination. When the robot arrives
at the destination, it opens the valve and waits for 1 second for the
object to completely \emph{detach} before starting the next
cycle. Note that the waiting times of $1$ second could be reduced.

All collision-free paths were planned \emph{online} using
\texttt{OpenRAVE}. Time-parameterization was also performed
\emph{online} subject to the grasp stability constraint in addition to
the robot's kinematic limits.

The experiment can be visualized in the second part of the
experimental video. Table~\ref{tab:comp-cost} reports our
findings. Pick-and-place cycles were less than $\SI{5}{sec}$ for all
objects. On average, the cycle time was $\SI{4.29}{sec}$, of which
$\SI{0.34}{sec}$ was for trajectory planning (including path planning
and time-parameterization), $\SI{1.95}{sec}$ was for robot motion, and
$\SI{2}{sec}$ was for waiting for the pump.

\section{Related works}
\label{sec:related-works}

In the robotic literature, many researchers have investigated the
``waiter problem'': a manipulator, equipped with a flat plate,
transports an object that is \emph{only placed} on the plate. This
problem is closely related to the one considered in this paper.  A
popular approach to achieve grasp stability guarantee, proposed
in~\cite{lertkultanon2014dynamic,Flores2013,Csorvasi2017}, is to
constrain the Zero-tilting Moment Point
(ZMP)~\cite{vukobratovic2004zero} of the object to its support area on
the flat plate. This clearly demonstrates the similarity between
object transportation and humanoid locomotion.  Similar to the
classical ZMP concept for humanoid locomotion, this approach has two
limitations: (i) it can only be applied if the object is on a flat
surface and (ii) there is no guarantee that the object does not slip
or twist~\cite{sardain2004forces}.  In a more recent
work~\cite{luo2017robust}, Luo and Hauser proposed to consider the
\emph{individual} contact forces explicitly, eliminating the use of
the ZMP and its limitations.  However, their formulation resulted in a
non-convex non-linear optimization problem that is computationally
demanding, taking several seconds to terminate.

In contrast, our proposed approach utilizes polyhedral computational
theory~\cite{fukuda1996double} to derive grasp stability conditions
that can account for general contact configuration (suction cup) and
at the same time, allow efficient computations.  Our use of polyhedral
computational theory is inspired by the development of the
Gravito-Inertia Wrench Cone in humanoid locomotion~\cite{Caron2017}.

The motion planning pipeline that we proposed follows the
Path-Velocity Decomposition principle~\cite{kant1986toward}. Planning
collision-free geometric path, the problem solved in the first stage,
is well-understood and can be solved
efficiently~\cite{LaValle2001,bordalba2018randomized,zucker2013chomp}. Solving
Time-Optimal Path Parameterization (TOPP), the problem solved in the
second stage, is also a classic problem in
robotics~\cite{bobrow1985time,verscheure2008practical,hauser2014,pham2014general}.
We solve TOPP using TOPP-RA, a recently introduced algorithm that is
both highly efficient and robust~\cite{Pham2017}. Interested readers
can refer to~\cite{Pham2017} for a more comprehensive review of the
literature on TOPP.

Lastly, many researchers have presented \emph{complete} robotic
pick-and-place systems~\cite{Correll2016,Morrison2017}, offering a
broader view of different modules such as motion planning, perception,
control and grasping. Regarding motion planning, these reported
systems commonly employ \emph{reactive strategies} that are based on
online visual feedback. In contrast, our approach fits entirely in the
conventional \emph{sense-plan-act strategy}. It is clear that with
sufficient knowledge of the external environment, our approach can
achieve a much higher level of performance; however, it remains to be
seen how will it perform in relatively uncertain environments.

\section{Conclusion}
\label{sec:conclusion}

We have proposed an approach for planning critically fast trajectories
for manipulators performing pick-and-place with suction cup. Before
execution, we identify the grasp stability constraint, the constraint
that the object must not fall from or slip or twist relatively to the
suction cup, as a system of linear inequalities. During execution, we
plan collision-free geometric paths for transporting objects and
retime these paths \emph{time-optimally} subject to the identified
grasp stability constraint using the TOPP-RA algorithm.

Experiments were conducted to assess the performance of the proposed
approach. The results suggest that the approach is capable of
producing high-quality trajectories: these trajectories can be
executed successfully without causing the object to fall or slip and
have duration that were close to the true time-optimal values.
Further, the proposed approach has a low computational cost, making it
suitable for online motion planning for industrial applications.

There are two limitations that we are actively investigating:
\begin{itemize}
\item Our approach requires exact knowledge of the object's inertial
  properties: center-of-mass position, mass and moment of inertia and
  the robot's geometry.  Yet, in practice, identification and
  modelling errors always exist; only approximations of objects'
  properties are available. This observation leads to two questions:
  1) what are the effects of identification errors on motion quality
  and 2) how to handle these errors;
\item The approximation step presented in
  Section~\ref{sec:strat-simpl-result} reduces computational time
  significantly as the size of the grasp stability constraint is much
  smaller.  However, its effects on motion quality is, in general, not
  completely understood.
\end{itemize}

\subsection*{Acknowledgment}

This work was partially supported by the Medium-Sized Centre funding
scheme (awarded by the National Research Foundation, Prime Minister's
Office, Singapore).

\bibliographystyle{IEEEtran}
\bibliography{library}

\end{document}